\documentclass[conference]{IEEEtran}
\ifCLASSINFOpdf
\else
\fi
\usepackage{graphicx}
\usepackage{amsmath}
\usepackage{amssymb}
\begin{document}
%
\title{A Comprehensive Study on Optimization Strategies for Gradient Descent In Deep Learning}

\author{\IEEEauthorblockN{Kaustubh Yadav}
\IEEEauthorblockA{School of Computer Science and Engineering, SCOPE\\
Vellore Institute of Technology\\
Vellore, Tamil Nadu \\
Email: kaustubh.q@gmail.com}
}


%


\maketitle

\begin{abstract}
One of the most important parts of Artificial Neural Networks is minimizing the loss functions which tells us how good or bad our model is. To minimize these losses we need to tune the weights and biases. Also to calculate the minimum value of a function we need gradient. And to update our weights we need gradient descent. But there are some problems with regular gradient descent ie. it is quite slow and not that accurate. This article aims to give an introduction to optimization strategies to gradient descent. In addition, we shall also discuss the architecture of these algorithms and further optimization of Neural Networks in general. 
\end{abstract}



%
\IEEEpeerreviewmaketitle

\section{Introduction}
Gradient Descent is one of the most important parts of Deep Learning used for optimizations of Neural Network. It is used to update weights so as to get the best result possible. Gradient Descent is a mathematical concept to calculate the minimum of a function, in case of deep learning it is calculated for minimizing the loss function or the cost functions by tuning the weights.  The gradient of a single neuron is dependent on the gradient of the previous neurons, and if you imagine a set fully connected layers how hard it can be to calculate this gradient and how computationally tedious it can be, this is the idea behind backpropagation which is itself derived from the chain rule. There are also problems regarding Backpropagation like the vanishing gradient and the exploding gradient problems which can be solved by weight initialization. With this article, we aim to provide a better understanding of gradient descent and its optimization and why this optimization is necessary. We shall also shed some light on other optimizing strategies that are being used in today’s Deep Learning Models. In Section 2, we will start with loss functions and backpropagation. In Section 3, we will discuss types of Gradient Descent and their optimizations in Section 4. Further, we will look at the performance of these optimizations on standard datasets in Section 5. With this, we will look at other optimization strategies and regularization in Section 6.

\section{Loss Functions}
Loss Functions are the measure of how good the neural network performs. In layman’s terms loss function is needed to calculate the deficit between the actual values and the predicted values by our network. Lower the value of the loss the better our network’s performance is and higher the value means that our predicted values are far from the actual values. And finally, our goal is to minimize this loss. Loss functions are of three types on the basis of what our network is known to perform, for example, Regression and classification. Let the input values are \( x_i \) and our outputs are \( y_i \), the loss function can be defined as:\\
\begin{equation}
L = \frac{1}{N}\sum_{i=1}^{N} L_i[f(x,W),y_i]+\lambda(R)
\end{equation}
Here $L$ is the average of the total loss incurred and $L_i$ is the loss function we will discuss in the following subsection. The $\lambda(R)$ term signifies regularization, which is another optimizing strategy that we will discuss in Section 6.\\

\subsection{Loss Functions for Regression Problems}
\subsubsection{Mean Square Error Loss}
Mean Square Error is one of the most common error function used in regression problems, it’s given by averaging the sum of the squared of the difference between the actual and the predicted values. This function is given by:\\
\begin{equation}
L = \frac{1}{N}\sum_{i=1}^{N}(y_i - y_{p_{i}} )^2
\end{equation}
Mean squared Error is ideal when the distribution of the target variable is Gaussian. The range of MSE lies from 0 to infinity, it can never be negative. Also, one important fact about MSE is that it gives a higher value of error when the deficit is higher, which means that model corrects itself when the error is larger.
\subsubsection{Mean Absloute Error Loss}
Although MSE works satisfactorily for Gaussian distribution, its performance decreases if our dataset has outliers, and continues to worsen if the outlies farther from the mean value. To solve this problem, MAE is used. It is calculated by averaging the absolute difference between the actual and predicted values. It is given by:
\begin{equation}
L = \frac{1}{N}\sum_{i=1}^{N}|y_i - y_{p_{i}} |
\end{equation}
MAE could be used if the distribution is not Gaussian. And as shown by Nie et. al \cite{nie2018investigation} outperforms any other loss functions if our dataset is noisy.
\subsubsection{Mean Square Log Error Loss}
MSLE is an intermediate between MAE and MSE, it gives almost the same priority to the values that are off by a larger magnitude and the values that are off by a smaller magnitude, which means that it doesn’t give too much priority to outliers. It is given by:
\begin{equation}
L = \frac{1}{N}\sum_{i=1}^{N}(log(y_i + 1) - log(y_{p_{i}} + 1) )^2
\end{equation}
But MSLE gives an asymmetric error curve because it gives a higher loss if our predictions have undershot the actual values. A padding of 1 is added because of the fact $log(0)$ is undefined.
\subsection{Loss Functions for Classification Problems}
The reason why we have separated both regression and classification is that we cannot use the losses we saw in the previous subsection as they won’t give a valid result. For example, we will take the case of MSE and we have to perform classification on a target value of 0 or 1. And our classifier will yield a value between 0 and 1. So if the value by the network is totally off ie. the actual value is 1 and the predicted value was 0 so by the  MSE equation the loss incurred will not be that large. Hence for classification problems, a different set of Loss Functions were defined. But this doesn’t mean we cannot use the aforementioned Loss functions in Classification problems, Ghosh et al. \cite{ghosh2017robust} proved that MAE actually performs better at classification than Cross-Entropy when tested on the MNIST dataset.
\subsubsection{Categorical Cross Entropy Loss}
Categorical Cross Entropy is used to measure loss between a softmax layer which contains the predicted class and one hot encoding which contains the true class. There are actually two variants of CCE. one is used for binary classification, this loss function is used when our target distribution is binary ie. there are only 2 output classes. This loss function actually a derived version of Cross-Entropy Loss with is another loss function for multi-class classification, but both of them have the same intuition. Binary Cross-Entropy Loss is given by:
\begin{equation}
L = \frac{1}{N}\sum_{i=1}^{N}y_i log(y_{p_{i}}) + (1 - y_i)log(1 - y_{p_{i}}) 
\end{equation}
Here $y_{p_{i}}$ is the probability of the class so its value ranges from 0 to 1 and $y_i$ is the true class and its value is either 1 or 0. The values of both the terms under summation will be negative because the log is negative from 0 to 1 hence we have a negative sign outside the summation. The reason why we use Binary cross-entropy is that it gives a very high loss for misclassification. For example, if our model gives a totally wrong result ie the actual value is 1 and the predicted value was 0. Because of the first log term, the Loss becomes infinity.  Although this property is crucial, this makes the loss function prone to outliers. A generalized approach to CCE was introduced by Zhang et. al \cite{zhao2015stochastic} where they used the robustness of MAE and the implicit weighting mechanism of the Cross-Entropy Loss which focuses on samples that are hard to learn. The other variant of Cross-Entropy is Multiclass Cross-Entropy, which is a broader generalization of Binary Cross-Entropy, so instead of two target classes in this can be and a number of target classes. It is given by:
\begin{equation}
L = \frac{1}{N}\sum_{i=1}^{N}(y_i)log(y_{p_{i}}) 
\end{equation}
As we discussed earlier Cross-Entropy has a drawback that it is quite prone to outliers but it does have a plus point in the sense that it focuses learning on "difficult examples" which mean it foucuses on the cases that are futher from the actual pediction than the labels that are closer to our samples, that's what makes it prone to outliers.
\subsubsection{Kullback-Leibler Divergence Loss}
KL divergence is defined as the deficit between the joint distribution and the product distribution of two variables in the case of Loss functions it is the probabilities of the true and the predicted labels. KL divergence is similar to entropy in the sense that they both have a logarithm term, but the difference is that Entropy is used with probabilities and KL Divergence is used with Probability Density Functions. A Probability function can be defined as:  
\begin{equation}
F(x)=P(a\leq x\leq b)=\int_{a}^{b}p(x)dx
\end{equation}
And we defined entropy by equation 6 but that was for discrete variables but PDF is continuous hence it can be defined as $p(x_i)\Delta x_i$ hence new entropy will be:
\begin{equation}
H(x) = -\sum_{i=1}^{n}p(x_i)\Delta  x_i . log(p(x_i).\Delta x_i)
\end{equation}
By properties of logarithms, we can write equation 8 as:
\begin{equation}
H(x) = -\sum_{i=1}^{n}p(x_i)log(p(x_i))\Delta  x_i - \bigg[\sum_{i=1}^{n}p(x_i)\Delta  x_i\bigg]log(\Delta  x_i)
\end{equation}
Taking the limits as $\Delta  x_i$ tends to 0
\begin{equation}
H(x) = \int_{-\infty}^{+\infty}p(x)log(p(x)).dx-\lim_{max(\Delta  x) \rightarrow o} log(\Delta  x_i)
\end{equation}
But there are problems with these equations that are they are not valid, a rule in applied math is that transcendental functions \cite{warren2018transcendental} and Probability density function do have units so the logarithms are invalid. Also as $\Delta  x_i$ approaches 0, $p(x)dx$ also approaches zero meaning that a little difference will give a huge value for $H(x)$. To solve this problem we use KL-Divergence, which is the difference between the joint distribution and the product distribution. Let $X$ be a variable with density $p(x)$ and $Y$ be a variable with density $q(x)$ the difference between joint distribution and product distribution is given by equation 11 which is also known as Relative Entropy or Kullback-Leibler Divergence.
\begin{equation}
D(X||Y)= -\int_{-\infty}^{+\infty}p(x)log(q(x)).dx + \int_{-\infty}^{+\infty}p(x)log(p(x)).dx
\end{equation}
Taking $p(x)$ as the common term
\begin{equation}
D(X||Y)= -\int_{-\infty}^{+\infty}p(x)[log(p(x))-log(q(x))].dx
\end{equation}
And finally by using the quotient rule for logarithms 
\begin{equation}
D(X||Y)= \int_{-\infty}^{+\infty}p(x).log\frac{p(x)}{q(x)}.dx
\end{equation}
So now we have a ratio under the logarithm hence this equation is totally valid. Also if $X=Y$ the value for $D(X||Y)$ becomes 0 and not infinty like in the case for equation 10. 
\subsection{Robustness Of Loss Functions}
The concept of robustness of loss functions \cite{huber2004robust} has been there for a long time and it is the first step in achieving optimization in deep learning. Here robustness of a Loss function means that it is not much influenced by larger errors. And a loss function with this capability can be universally applied on any dataset. For example, MAE is more robust to outliers than MSE, due to the fact that the difference is squared in case of MSE hence the outliers have a higher loss than the loss in MAE. But when we are optimizing our losses with gradient descent it is the derivative that matters. A larger value for the loss doesn’t mean that that the derivate would large as well. The derivative depends on the loss function that is in use. A larger value for the loss(outlier) will have a negligible derivative. In addition to this, the entries in our dataset or examples can be classified into two types- easy and hard examples. Easy examples are easier to learn and the model focuses on these sorts of examples which usually have a lower error, but these examples are known to slow convergence ie. during Gradient Descent these examples will take a lot of time so as to achieve optimal results. The other type is hard examples, these examples are known to accelerate the convergence to achieve optimal results faster. Hence Hard Example mining \cite{shrivastava2016training} is a big part of the research on optimization. Hence when someone says more the data better the performance it is quite an oversimplification, in the end, these hard examples are the key to achieve the results we require faster. But when it comes to hard or easy examples we ourselves cannot differentiate between them and nor can our model. It is after some training we can be a little certain with the difference between them. Even then robustness matters that is our model might start to focus on noisy data, this can be solved by some variations of Stochastic Gradient Descent as discussed in Section 4. This also brings up the concept of learning, as with human leaning patterns we can apply the same patterns in machine learning, some of the popular learning patterns are Curriculum Learning and Self Paced Learning. 
\subsubsection{Curriculum Learning}
 This method is quite inspired by how humans learn. It was introduced into machine learning by Bengio et al. \cite{bengio2009curriculum} as a concept for attaining convergence faster and finding a local minima for non- convex functions. The intuition behind CL is to give the machine easier examples and then consequently give harder examples when it’s ready. This can also be observed while training a CNN, at every layer, our model learns a new level of abstraction. 
 \subsubsection{Self-Paced Learing}
 The motivation behind this method is quite similar to Curriculum Learning in the sense that we initially cannot give our model everything and expect it to learn the parameters, instead we start with easier examples and then move to harder ones. If we have a dataset of animals and we need the model to classify it into its corresponding label, the measure of easiness is totally ambiguous, furthermore, if the dataset has manually boxed images it’s not conclusive that the example is easy or hard. Technically, the algorithm for self-paced learning takes inspiration from the Concave Convex Procedure (CCCP)\cite{yuille2003concave} which is a solution to Latent SVM\cite{kumar2010self}. For a binary linear SVM classifier, the objective is to get the maximum distance between two hyperplanes for the corresponding classes, which is given by:
 \begin{equation}
 max \bigg[(\vec{x}_+ - \vec{x}_-). \frac{\vec{w}}{||w||}\bigg] = min \frac{1}{2} ||\vec{w}||^2
 \end{equation}
 And the Decision Rule is given by: 
 \begin{equation}
 y_i(\vec{w}x_i+b)\geq 1-\xi_i 
 \end{equation}
Here $\vec{x}_+$ and $\vec{x}_-$ are the two classes and $\vec{w}$ is the weight vector and $y_i$ is +1 for positive samples and -1 for negative samples. The minimization can be calculated by taking the gradient with respect to weight and bias. But we cannot get a perfect hyperplane always because the data might be inseparable hence there are two methods for a generalized result, the first one is transforming a vector using a kernel function\cite{amari1999improving}, but this virtually increases the complexity of the function, hence we use the second method which is by using a slack variable $\xi$, which is a penalty of how much we are not able to satisfy our Decision Rule. Now Latent SVM works on the premise that we have to find a value of the hidden variable that should be consistent with our labels should be better than any other pairs of ground truth and hidden variables, if not so we will add a penalty(slack variable) which we can further minimize. So in the following equations, $h_i$ are the set of hidden variables on a dataset $ D={{x_i,y_i}}$ and $\psi (x,y,h)$ is our feature space.
\begin{equation}
max_{h_i} w^T\psi (x_i,y_i,h_i) - w^T\psi (x_i,y,h) \geq \nabla(y_i,y,h)-\xi_i
\end{equation}
But this is still a Non-Convex Solution, hence we use CCCP, which first initializes $w_0$ randomly and then updates:
\begin{equation}
\frac{1}{2}||\vec{w}||^2 + C \sum_{i=1}^{n} \xi_i
\end{equation}
And finally update $w_{t+1}$ by solving convex problem, $min||w||^2 + C \sum_{i}^{}\xi_i$ and equation 16.  But this still does’nt solve the problem of harder and easier example hence Self Paced Leaning introduces a new variable to equation 17, $v_i \in {0,1}$ where 1 defines easier examples and 0 defines that we will not be bothered by that example for now. Making the equation as $min||w||^2 + C \sum_{i}^{}v_i\xi_i$. But this will be trivial as we can put all $v_i$ to be 0. Hence adding penalty, which doesn’t allow all the values to be 0 and for every value 0 the penalty will be higher and higher. Making equation 17 as: 
\begin{equation}
min||w||^2 + C \sum_{i}^{}v_i\xi_i-\sum_{i}^{} \frac{v_i}{K}
\end{equation}
Where K is the self-paced learing rate weight, initially K is large hence hard examples are excluded and sequentially decreased to inclued hard examples. 
\section{discrepancies in Non-Convex Optimizations}
In this section, we will look into the problems that arise while optimizing non-convex problems, the reason for this is that when we look at our loss functions against the weights, through every round of backpropagation \cite{rumelhart1986learning} we are looking for the set of weights that minimize our loss function. Hence here the word optimization really means that we have to minimize the cost function which is non-convex. The best way to minimize any function is to compute the gradients and move along the direction of its decrease. But this is quite an overstatement when it comes to saddle points and local minima or even local maxima as the gradient at these points is zero. Although gradient descent performs well in a convex setting as it converges in $O(1/\epsilon)$ iterations and even better in stong convex as the convergence rate is poly-logarithmic ie. it is almost not dependant on the dimensions. But deep learning problems are highly non-convex, with saddle points, local minima, low gradient regions and we cannot expect regular gradient descent to converge to the global minimum and not get stuck at these sub-optimal regions. Fortunately, we can sort of rule out local minima as they give quite satisfactory results, this was proved for many cases including dictionary learning \cite{sun2016complete}, matrix completion \cite{ge2016matrix}, tensor decomposition \cite{ge2015escaping}, and some cases of deep learning \cite{kawaguchi2016deep}. Generally, for deep learning, a local minimum is almost as good as global minima \cite{choromanska2015loss}. But this still doesn’t address saddle points and bad local minima.
\subsection{Saddle Points}
When we use gradient descent on these non-convex problems, it cannot distinguish between a saddle point and a local minimum as the gradient is 0 in both the cases. Hence Gradient Descent gets stuck maybe indefinitely or as long making training time infeasible, and we get a sub-optimal result. Prior works on escaping saddle points involved firstly getting to a stationary point and if it is a local minimum we stop but if it is a saddle point we have to escape from it. Escaping from the saddle point involves following the negative eigenvector of the Hessian when the gradient is 0($\nabla f(x)=0$) and the hessian is positive semidefinate ($\nabla^2 f(x)\succeq0$). In a 2-D approach, if we look at the gradient flow around a saddle point we have a direction where the gradient is flowing towards the saddle point and a direction where the gradient is flowing away from the saddle point, hence if we start at a point which is not on the central line, we can escape the saddle point using Stochastic Gradient Descent in $O(1/\epsilon^4)$ rounds \cite{ge2015escaping} and $O(1/\epsilon^2)$ in full Gradient Descent \cite{jin2017accelerated}.But this result is highly constrained to first-order saddle points where a saddle point is nth order:
\begin{equation}
f(x’) \geq f(x)- O(\|x-x’\|^{n+1})
\end{equation}
Hence a normal saddle point ie. $z=x^2 - y^2$ is a first-order saddle point. But there are a lot of different types of saddle points specifically of higher orders, for example, a monkey saddle which is a second-order saddle point given by($y={x_1}^3 - 3{x_2}^2x_1 $) and hyperbolic paraboloid which is also a first-order saddle point. Optimality conditions are defined for the first-order saddle point that if the gradient is not zero we move in the direction of the negative gradient and if the Hessian is not positive semidefinite we follow the direction of the negative eigenvector of the Hessian, but these conditions don’t include second-order saddle points. Hence an extension by Anandkumar and Ge \cite{anandkumar2016efficient} where they included a special constraint that is any vector should be negative in the null space of the Hessian and the third derivative should be zero in this direction which is given by:
\begin{equation}
\forall u, [\nabla^2 f(x)]u=0;[\nabla^3 f(x)](u,u,u)=0
\end{equation}
which is not true for the Monkey Saddle and hence this can also be an optimality condition when the second term is not equal to zero and hence in that case we pick a random $u$ in the null space of the hessian and move in the direction of either $u$ or $-u$. In case of deep learning problems, it is quite rare to see higher-order saddle points so the final optimality conditions really sums it all up.
\\
Although the methods mentioned above are quite robust, they do have two limitations, one that we have to be in the saddle point and then switch to a strategy to escape it which inherently leads to longer training times, and secondly, we need 3rd order information which is hard to compute. Hence a method by Zeyuan Allen-Zhu \cite{allen2018natasha} we just “swing by” a saddle point, which means that we have knowledge on when our point will get stuck in a saddle point and with that knowledge we perturb so as to avoid the saddle points. A saddle point is called a $\delta$ strict saddle point if eigenvalue is below $-\delta$ hence if we have any point y and $\|y-x\| \leq \gamma$ where $\gamma < \delta$ due  second order Lipschitz conditions some eigenvalue of $\nabla^2 f(y)$ are below $-\gamma$. Hence if we have multiply it with a vector($v$) and a transpose we can almost be sure if that vector is close to the saddle point. Also now we can move in the direction of the negative or positive of this vector ie. $y'=y\pm \delta v$ which will steer us away from a saddle point.

\subsection{Bad Local Minima}

This might seem a bit counter argumentative to the statement that local minima are as good as global minima but when we see the holistic picture, we have to avoid those local optimum that give sub-optimal results and our loss functions have a lot of local minima. Some studies have shown that by adding one neuron can actually eliminate all the bad local minima, the theoretical proof has been given in multiple studies \cite{liang2018adding} and \cite{kawaguchi2019elimination}but was initially proved by Auer et al. \cite{auer1996exponentially} for square loss using logistic activation function for the extra neuron. The original study changes the loss function to a different form and then gives the evaluations based on that transformation, in general, we define our loss function as $L(f(x,W),y_i)$ but now we have a separate function $l$ which takes one argument $l(-y_if_0(x,W))$ where $l$ is assumed to be twice differentiable and non decreasing and all critical points to be the global minima, one another strong assumption is that the dataset is serializable that is the model can classify the given parameters correctly which is not a common attribute when it comes to datasets in general. It was observed by the original authors that the new loss functions can be visualized as a polynomial hinge loss. Hence we modify our loss function to become:

\begin{equation}
L =\sum_{i} l(-y_i f(x_i))+\frac{\lambda a^2}{2}
\end{equation}

Where $\lambda$ is the regularization term and we change our model output to $f(x,W)=f_0(x,W)+f_e(x_i)$ where $f_e(x_i)= a*exp(x_i W^T + b)$ and $a$ is the scale and $b$ is the bias. After making a new loss function, usually, we calculate the derivatives respect to weights, but in order to show that effect of the extra neuron vanishes when we reach a global minimum, derivatives with respect to scale and bias are needed, which will be same but the derivative with respect to the scale will have an extra $\lambda*a$ term and finally to find the minimum we equate both the deviates to zero. Now just by subtracting both the derivatives and multiplying $a$ on both sides we e $\lambda* a^2=0$ which says with a non-zero $\lambda$, $a$ has to be zero and hence for the global minimum the extra neuron becomes inactive. 

\section{Gradient Descent and it's variants}
The idea of gradient descent is a derivative of the descent methods that were introduced about 150  years ago, hence before going into what gradient descent is we will start with the initial idea, which is the steepest descent and was given for strongly convex functions. A function $f$ is convex if  \emph{dom}$f$ is a convex set and it satisfies:

\begin{equation}
f(\theta x+(1- \theta)y) \leq \theta f(x) + (1-\theta)f(y), \forall x,y \in \emph 0\leq \theta \leq 1
\end{equation}

And using the first-order conditions for convexity the first-order Taylor approximation is always a global underestimator of the function ie. $f(y)\geq f(x)+ \nabla f(x)^T (y-x)$. Also, we can define a function as $\alpha$ strong convex if it satisfies:
\begin{equation}
f(y)\geq f(x)+ \nabla f(x)^T (y-x)+ \frac{\alpha}{2}\|x-y\|^2 
\end{equation}
Where the quadratic implies that there exists a lower bound to the growth of the function which in turn means that strong convex growth is strictly greater than linear growth, also strong convexity is a dual-assumption to Lipschitz smoothness. Also let $q^*$ be the optimum value for the function using Polyak-Lojasiewicz inequality we can also derive $f(x)-q^* \leq \frac{1}{2\alpha}\|\nabla f(x)\|^2$ which also calls for a stopping criterion as the gradient tends to zero. To optimize smooth convex problems descents methods are the key and generally, they can be defined as:
\begin{equation}
x^{k+1} =x^k - \eta \Delta x^k, f(x^{k+1}) <f(x^k)
\end{equation}
Where $\Delta x^k$ is called the normalized step direction which is usually a vector and $\eta$ signifies the step size which is always positive.  Hence general descent methods involve the calculation of the step direction $\Delta x$ and then doing a ray search \cite{nocedal2006numerical} to for the step size $\eta$ and then updating equation 24, until we satisfy the stop criterion. The reason we used the word ray search in place of line search is that we only look for positive values of $\eta$. A basic line search or exact line search works on the premise $\eta = argmin_{\eta>0} f(x +\eta \Delta x)$ which really means that we are choosing $\eta$ to achieve the minimum of the plot. Another variant of line search is the backtracking line search which performs a bit better than exact line search \cite{yuan2010line}. In backtracking line search, we start with $\eta = 1$ and we update $\eta =\beta \eta$ until the following condition is satisfied $f(x + \eta \Delta x) < f(x) + \alpha \eta \Delta f(x)^T \nabla x$ where the parameters $\alpha \in (0,0.5)$ and $\beta \in (0,1)$ also this condition is invalid if the $\alpha \geq 1 $ due to preservation of convexity and as the first-order taylor approximation is a global underestimator of the function.
When we observe the equation for steepest descent, there is an anomaly ie. it is not normalized which means in order to multiply with the step size we have to get them into the same norm. As our intuition, $Delta x$ should have the gradient with respect to $x$ hence normalized step direction can be given by $\Delta x_n = argmin{\| \nabla f(x)^T v \|}$ where $v$ is the direction derivative. General descent methods use the un-normalized step direction that is $\Delta x = \| \nabla f(x)\| \Delta x_n$. Also, Gradient Descent is actually steepest descent in the Euclidian Norm or the 2-norm. Hence now the only thing changed from the definition of Steepest descent is that Gradient Descent has a step direction negative to the gradient with respect to $x$. Also, with fixed step size, the convergence rate is of $O(1/k)$ where $k$ is the number of iterations and is same even with backtracking line search but as $\eta$ changes in every iteration, we replace $\eta$ with $\eta_{min} = min\{1,\beta /L\}$ where $L$ is the Lipschitz Constant and is always greater than 0.  

\begin{figure}[h]
	\centering
	\includegraphics[width= \columnwidth]{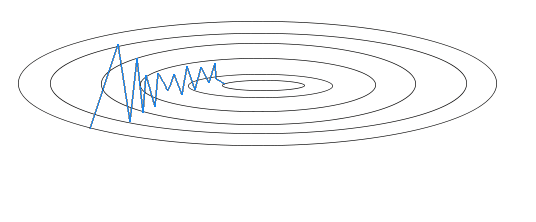}
	\label{Figure 1}
	\caption{Sub-level sets of a function $f(x,y)=x^2+by^2$ or $f(x)=\frac{1}{2} x^T A x$}

\end{figure}

As we can see it is the norm that matters, it is only useful if we can change the norm for better convergence. As in Figure 1 if the sublevel sets are oval and narrow, here when we apply the steepest descent it works very poorly, but if the sub-level sets are isotropic, ie. almost spherical then the direction of the negative gradient will either be on the minimum or extremely close to it. Also when we get close to the minimum of the function the gradient tends to zero, but that doesn’t mean we are at the minimum, so as that happens we switch to the second-order Taylor approximation which can be written by appending  $\frac{1}{2}(y-x)^T \nabla^2 f(x)(y-x)$ to the first-order approximation. Hence near the optimum, the sub-level sets look like ellipsoids due to the hessian almost similar to Figure 1, so initially, if we have some intuition about the hessian we can change the norm in accordance to the hessian and then we can apply steepest descent, this is called Newton’s step. So by the newton’s step we change the step direction $\Delta x$ as $\Delta x_{ns} = -\nabla^2f(x)^{-1} \nabla f(x)$ or else another interpretation would be that we take the second-order approximation and minimize that or else we need to find a $v$ such the $\nabla f(x+v)$ is zero. After changing the norm and using newton’s step as the new step direction we have to change the stopping criterion as well, which is described as Newton’s decrement and is given by $\lambda (x)=(\nabla f(x)^T \nabla^2 f(x)^{-1} \nabla f(x))^{1/2}$ which is a measure of closeness to the optimum similar to the stopping criterion of the Descent methods, hence the best approximation ie. $f(x)-q^*$ is $\lambda^2 / 2$. So in conclusion, Newton’s method requires to calculate Newton’s step $\Delta x_ns$ and Newton’s decrement $\lambda^2 $ and update $x^{(k+1)} := x^{(k)} + \eta \nabla x_ns$ until the stopping criterion that is $\lambda^2 /2 \leq \epsilon$
Although gradient descent is a derivative of steepest descent the assumption that it performs better than other descent methods is totally arbitrary because if we have a norm which makes our function align with which geometry of the sublevel sets, the convergence would be extremely fast. Also, Newton’s method is more advantageous in the sense that if we scale it or maybe take a different norm, it will converge almost in the same number of steps whereas gradient descent will lead to a drastic impact on changing the norm.  

But in deep learning and machine learning, we generally tend to use gradient descent, hence going into depth with more descent methods and second-order methods are out of the scope of this paper. As gradient descent is a quite widely researched area, there are 3 types of gradient descent variants that differ on the amount of data required of the calculation of gradient for the loss functions. 
\subsection{Batch Gradient Descent}
Also known as vanilla gradient descent, is the first intuitive step in optimizing our cost functions, as this classification was based on the amount of data, in batch gradient descent entire dataset is used for the gradient update. 

\begin{equation}
x^{k+1} := x^k -\eta \nabla_{:n} f(x_{:n})
\end{equation}

where $n$ is the number of entries in the dataset. Hence to make one gradient update the gradient for every parameter is calculated. For large datasets, this is an extremely tedious process and might take hours just for one update, and for extremely large datasets, it doesn’t converge at all. But it is extremely accurate for small size problems in the sense that it will converge to a global minimum for convex and a local minimum for a non-convex setting.

\subsection{Mini-Batch Gradient Descent}
As the name states, mini-batch gradient descent is a truncated version of vanilla gradient descent, what that means is that in mini-batch gradient descent we divide our parameters and the labels into smaller sets and with each iteration calculate the gradients with respect to one of the mini-batches.

\begin{equation}
x^{k+1} := x^k -\eta \nabla_{i:i+n} f(x_{i:i+n})
\end{equation}

Hence instead of calculating the gradient for the entire dataset, we calculate the gradient for only a mini-batch. The convergence plot is a little different from batch gradient descent in the sense that in batch gradient descent the plot with respect to cost and iterations only goes downward but in the case of mini-batch the plot is noisier because in every iteration we are taking a new batch for our descent. Now another ambiguity that arises is the size of the mini-batch if our training data is small that is, 3000-4000 entries it’s recommended to use Batch gradient descent because even if we divide it into mini-batches the convergence wouldn’t have a that significant improvement if any, although in case of large datasets mini-batches it’s recommended to use mini-batches of the powers of two \cite{li2014efficient}, also larger the batch size small the gradient improvement hence there is a trade-off between the batch size and the convergence.

\subsection{Stochastic Gradient Descent}
Stochastic Gradient Descent is essentially a condensed form of mini-batch gradient descent, here the size of the mini-batch is 1.

\begin{equation}
x^{k+1} := x^k -\eta \nabla_{i} f(x_{i}); i \in rand \{1,2,..n\}
\end{equation}

 Hence during a gradient update, we only need to get the gradient of one training example, which is usually picked at random. Now if we compare Batch gradient descent with SGD, it is quite a jump in the sense if we have 1 million training examples, Batch gradient descent has to compute 1 million gradients each iteration and if we have 1000 iterations, so we have to calculate gradients 10 billion times whereas in SGD we only need one gradient per iteration hence SGD is 1 million times faster than Batch Gradient Descent, on the premise both reach convergence. Another feature for SGD is that it is very sensitive to the learning rate,$\eta$ \cite{musso2020stochastic}, and as the leaning rate increases so does the variance. The reason why SGD is common in machine learning is that, as stated earlier we are not looking for the perfect solution in the sense that the optimum should most definitely be a global solution, our aim is to get an optimum that performs well on unseen data, so in SGD essentially we get a considerable drop in the cost initially and that is what we need, so with early-stopping we get the result that we need \cite{caruana2001overfitting}.

\section{Optimization of Gradient Descent}

Now, when we talk about optimization there’s a lot that can be done as gradient descent itself has a lot of moving parts. Although we did touch upon some optimizations that do give better results in the early parts of the previous section but those methods involve the use of second-order methods which are not easy to compute talking from a deep learning perspective. Also, the problems discussed like bad local minima and saddle points arise here as well and hence here we shall provide an GD approach to those problems rather than second-order methods.

\subsection{Gradient with Momentum}
As we saw in Section 3, saddle points are mostly responsible for the sub-optimal results when it comes to optimization and even if the surface for our loss function is flat, gradient descent still runs into problems pertaining to the low gradients and hence slow movements. Momentum with Gradient Descent\cite{qian1999momentum} \cite{rumelhart1986learning} is based on the intuition we have from motion that is an object in motion will continue in the same direction until an equivalent force is applied in the opposite direction, here we care about the first part of the previous statement. Now in the sense of optimization, we move in the direction of the negative gradient and continue to update in that direction. Also, when we see convergence on our sublevel sets in Figure 1, we can see that it’s a zig-zag motion, if the step size is high then we might overshoot out of the sub-level sets and then even convergence towards the minimum is not guaranteed. So it is quite obvious that along the y-axis should be as minimum as possible and along the x-axis should have considerable but constrained updates, this can be solved by momentum. As shown earlier, gradient descent can be represented as
 \begin{equation}
 x_{k+1}:=x_k - \eta Z_k
 \end{equation}
 
 where $Z^k = \nabla f_k $ so with every step we go in the direction of the negative gradient, after introducing the momentum term we have a memory of the previous update ie. $k-1$, so
  \begin{equation}
  Z_k =\nabla f_k + \beta Z_{k-1}
  \end{equation}

  where $\beta$ is essentially the momentum parameter and $Z_{k-1}$ holds the information of the previous step direction. Now, by using the same example from Figure 1, the gradient is actually $Ax$ so by replacing $\nabla f_k$ in equation 29 and replacing $k$ as $k+1$ we get a system of equations,
  
  \begin{equation}
  x_{k+1} := x_k -\eta Z_k;
  Z_{k+1} - AX_{k+1}=\beta Z_k
  \end{equation}
  which can be written in a matrix form as,

\begin{equation}
\begin{pmatrix}
1 & 0 \\ -A & 1
\end{pmatrix}
\begin{pmatrix}
x_{k+1} \\ Z_{k+1}
\end{pmatrix}
=
\begin{pmatrix}
1 & \eta \\ 0 & \beta
\end{pmatrix}
\begin{pmatrix}
x_k \\ Z_k
\end{pmatrix}
\end{equation}
And finally, at every step, we have to compute the matrix
\begin{equation}
\begin{pmatrix}
c_{k+1} \\ d_{k+1}
\end{pmatrix}
=
\begin{pmatrix}
1 & -\eta \\ \lambda & \beta -\eta \lambda
\end{pmatrix}
\begin{pmatrix}
c_k \\ d_k
\end{pmatrix}
\end{equation}

Which depends on both $\beta$ and $\eta$, hence we have to minimize that matrix, which in turn is dependent on the eigenvalues, so we minimize over a whole range of eigenvalues. Let $m$ and $M$ be the upper bound and the lower bound on our eigenvalues, hence we have to choose $\beta$ and $\eta$ so as to minimize that eigenvalues for a whole range of $\lambda$ between $m$ and $M$ and it has been proven\cite{rumelhart1986learning}for  the optimality points for $\eta$ and $\beta$,

\begin{equation}
\eta_{optimal} = \bigg(\frac{2}{\sqrt{M} + \sqrt{m}}   \bigg)^2
\end{equation}

\begin{equation}
\beta_{optimal} = \bigg(\frac{\sqrt{M} - \sqrt{m}}{\sqrt{M} + \sqrt{m}}   \bigg)^2
\end{equation}
 Usually, $\beta$ is set at 0.9. Now, what the momentum term does is that as it records the previous step direction it moves in that direction, by the magnitude of $\beta$ hence the frequency of the oscillations drops and the convergence is accelerated. 
 \subsection{Nesterov's Accelerated Gradient}
After the introduction of the momentum term, the frequency oscillations were considerably reduced but there is always a scope for more. Secondly, the momentum term works well on the premise that our data is free from any sort of considerable outliers which means that while updating if we have an outlier and the gradient at that points in a direction opposite to the minima, every consequent update will have point to that direction at least for the foreseeable updates, increasing the convergence time. Also, this phenomenon is quite common in a stochastic setting as the update only looks at one entry while update. Nesterov’s Accelerated Gradient(NAG)\cite{nesterov1983method} introduced the concept of a lookahead term in extension with classical momentum. So instead of first calculating the gradient and then moving in that direction and doing a final update with regards to the previous step, we can just move with regards to the previously recorded step and then calculate the gradient at that point and then move accordingly. So we are calculating the gradient essentially at a different point$y_k$ than the initial starting point($x_k$). Firstly as we need to compute the gradient for the sake of simplicity we can define a function$g(x)$ as $g(x)=x-\eta \nabla f(x)$ which is same as computing the gradient and updating in its direction of decrease. So as we have to compute the gradient at a different point we can write the function $g(x)$ as $g(y_k)$

\begin{equation}
x_k=g(y_k) = y_k-\eta \nabla f(y_k)
\end{equation}

Also, initially $y_1=x_0$. After the calculation of gradient at the new point $y_k$ we have to make a correction, that is we have to change the direction of the previous momentum step which is given by,

\begin{equation}
y_{k+1} = x_k -\frac{t_k-1}{t_{k+1}}(x_k -x{k+1})
\end{equation}

Here, $t_k-1/t_{k+1}$ is essentially known as Nesterov’s coefficient also represented by $\kappa$. And $t_{k+1}$ is defined by $t_{k+1}=\frac{1+\sqrt{1+4{t_k}^2}}{2}$ and initially $t_1 =1$. Now, these set of equations can guarantee fewer oscillations as compared with Classical Momentum, as a matter of fact, the convergence rate is $O(1/k^2)$(k-iterations) in Nesterov’s Momentum whereas the convergence rate in Gradient descent is $O(1/k)$ hence there is a significant boost in convergence rates. But, it still doesn’t address the problems with SGD as mentioned earlier and as a matter of fact, the convergence rate drops to $O(1/\sqrt{k})$ while using Accelerated Gradient with SGD, which is even worse than normal Gradient Descent. So we can actually improve the performance of SGD with Nesterov’s Momentum to $O(1/k)$ quite easily, by using Variance Reduction with Stochastic Gradient Descent, also known as SVRG\cite{johnson2013accelerating}. But getting the convergence rate of $O(1/k^2)$ is still under research. Although a method known as Katyusha’s Momentum\cite{allen2017katyusha} was introduced for getting a convergence rate of $O(1/k^2)$ and it was an extension to the concepts of Nesterov’s Momentum. Instead of computing the gradients and taking the step with respect to prior momentum, we compute the next point with our momentum term form the previous time step but instead of making a definitive update, we compute the midpoint between the new and the initial point and do the further updates in the same manner from the midpoints. And this method gave a convergence rate congruent to NAG but was also dependent on the number of examples we are using in total.

\subsection{Adaptive Gradient Algorithm (AdaGrad)}
Although we have seen a good convergence rate from Nesterov’s Momentum and Katyusha’s Momentum it’s important to note that these convergence rates are not universal and may vary tremendously under certain cases. Another flaw that is quite common in the aforementioned methods is overshooting, although quite reduced with Nesterov’s and Katyusha’s Momentum but still prevalent. Adaptive Gradient Algorithm or AdaGrad\cite{duchi2011adaptive} solved this discrepancy by introducing the concepts of adaptive step size, which when tested performed really well on large scale and distributed networks. They also work well for sparse data like word embeddings\cite{pennington2014glove} also due to the fact step size changes for every iteration. The intuition behind the adaptive gradient is that when we start the sum of the prior gradients will be smaller compared to the sum considered after simultaneous iterations. As we just make a change in the step size which is a scalar quantity the update equation is quite similar to that of Gradient Descent.

\begin{equation}
x_{k+1} =x_k - \eta{'} \Delta x_k
\end{equation}
  where $\eta'$ indicates the new step size for the $k$th iteration, so $\eta'$ changes with every iteration, although as a hyperparameter we have to decide the initial step size which is usually kept at $0.01$, which gets decrimented later on. Now, $\eta’$ is defined as,
 
 \begin{equation}
 \eta’ = \frac{\eta}{\sqrt{diag(G_i)+\epsilon I}}
 \end{equation}
 
 where $\epsilon$ is a small positive scalar quantity which is appended to make sure that $\eta’$ is constrained even if the other term is zero. And $G_i$ is the matrix that stores the sum of the products of the gradients till the $i$th iteration represented as $G_i=\sum_{1}^{i}{\big( \nabla {x_k}^{(i)}\big)}^2$. Hence the diagonal of the matrix $G$ will hold the elements that we actually need and taking the square root of the entire matrix isn’t actually required. As we are multiplying the gradient by itself, if the value for $diag(G_i)$ is large during the later iterations of gradient descent hence the step size automatically drops down. Hence, a small gradient pertains to bigger step size and a bigger gradient pertains to smaller step size. But if the initial gradients are large then the step sizes for the rest of the iterations will be smaller and smaller this can be corrected by using a higher learning rate than $0.01$. But there are problems with this method too firstly, it doesn’t perform well in a non-convex or dense setting, AdaGrad can get easily stuck in saddle points and to get out we are only dependent on the value of $\epsilon$ which will lead to increased training time. Also if the value for the sum of gradient becomes extremely large, it will eventually make the learning rate so small that our model won’t learn anything after that point on, that is why AdaGrad is usually used in cases of sparse data.
 \subsection{AdaDelta}
 
Adadelta\cite{zeiler2012adadelta} was introduced as an extension of AdaGrad, improving on the drawbacks it had which are the accumulation of gradients leading to preponed stoppage of training and manual setting of the initial learning rate. The first constraint introduced in the paper was limiting the window size of the previous gradients, hence instead of storing the $t$ iterations, we keep information for $w$ iterations. By doing the denominator term in AdaGrad will not reach infinity and updates will be done on the basis of recent gradient information which in turn doesn’t terminate learning prematurely. Also storing the previous, squares of the gradient is really inefficient as we have to define a separate matrix and use and update that it each step, instead we can compute the average of the square of gradients or expectation given by,

\begin{equation}
\mathbb{E}[g^2]_t = \rho \mathbb{E}[g^2]_{t-1} + (1-\rho){g^2}_t
\end{equation}
here  $\rho$ represents the decay constant and has a similar function to $\beta$ in classical momentum. The value of $\rho$ is usually kept at $0.9$, so by Equation 39, we can intuitively say that expectation at a point $t$ is influenced both by the expectation of the square of gradients at the previous point and square of the gradients at the point more so in the previous steps.  As described in AdaGrad we require the square root of the squares of the gradients, we get the square root of $\mathbb{E}[g^2]_t$ which is also Root Mean Square(RMS). Also, we still need a small positive number $\epsilon$ to escape regions of low gradients, hence $RMS[g]_t= \sqrt{\mathbb{E}[g^2]_t + \epsilon}$ and the new learning rate becomes,

\begin{equation}
\eta’ = \frac{\eta}{RMS[g]_t}
\end{equation}

The original paper also pointed out another discrepancy which was the units of the parameter update term, $\eta \nabla x_k$ which is proportional to the units of $1/x$ on the premise that the cost function doesn’t have any units, but if we compare these finding with second-order methods such as Newton’s method the units for the update will be directly proportional to $x$ using the formula $\Delta x_{ns} = -\nabla^2f(x)^{-1} \nabla f(x)$ where $\Delta x_{ns}$ is the update parameter for Newton’s step as explained in Section 4. Hence with a little rearrangement, we can show that the inverse of the Hessian is equal to $\frac{\Delta x_{ns}}{\nabla f(x)}$, also here the parameter $\Delta x$ is given by $\eta’ \nabla x_t $ hence as the RMS information is represented as the denominator hence $\Delta x_t$ is in the numerator, which needs to be calculated for the time step $t$ but by assuming Lipchitz smoothness we can approximate $\Delta x_t$ as $RMS[\Delta x]_{t-1}$ so the parameter update becomes,

\begin{equation}
\Delta x_t = \frac{RMS[\Delta]_{t-1}}{RMS[g]_t}g_t
\end{equation}
And the final update is, $x_{t+1}=x_t-\Delta x_t$. Now the introduction of $RMS[\Delta x]_{t-1}$, which essentially lags behind one-time step, made an improvement ie. it made this model robust to sudden changes which can stall the learning rates. Also, this method is analogous to Momentum as the RMS information holds the information of the previous time steps as momentum holds the history of the previous time steps
\\
Another novel idea which is quite similar to Adadelta is RMSprop which also pursued the same goal of having an adaptive learning rate and intuitively is very similar to Adadelta as it is also governed by 
\begin{equation}
\begin{split}
\mathbb{E}[g^2]_t = \gamma \mathbb{E}[g^2]_{t-1} + (1-\gamma){g^2}_t\\
x_{t+1}=x_t- \eta’ g_t
\end{split}
\end{equation}

Where $\eta’$ is the same as the one defined in Equation 40. The only difference is this method doesn’t involve the RMS values of the previous time step, but the rest is the same. Performance-wise, there’s is not a monumental difference but Adadelta performs better in cases of highly non-convex loss functions\cite{zhou2018convergence} but Adadelta still outperforms SGD, momentum and Adagrad under certain tasks\cite{zeiler2012adadelta}.

 \subsection{Adam}
 Adam \cite{kingma2014adam} is yet another first-order optimization for gradient descent and in some sense is the union of RMSprop and Momentum. Hence it retains the information for both the square of the past gradient as in RMSprop and also the gradient information as in Momentum. Although Adam works with adaptive learning rates, it has 2 parameters that have to be tuned manually $\beta_1$ and $\beta_2$ where $\beta_1,\beta_2 \in [0,1)$, which are decay rates that control the expectations of gradients and the sum of squares of gradients. Hence first we define $m_t$ which is defined as the first movement and involves the gradient information of the previous time steps, this variable is intuitively similar to the one in Equation 29 given by,

 \begin{equation}
 m_t=\beta_1 m_{t-1} + (1-\beta_1) g_t
 \end{equation}
 
 and $g_t$ is the gradient of the objective function at time $t$. It’s obvious that this term is quite similar to the momentum update as we are keeping a record of the previous gradients. The second variable is $v_t$ which records the sum of the squares of gradient till the time step $t$ similar to Equation 42 in RMSprop given by,
 
 \begin{equation}
 v_t=\beta_2 v_{t-1} + (1-\beta_2) {g_t}^2
 \end{equation}
 
 here ${g_t}^2$ stores the sum of gradient till time step $t$. Now if the gradient update is done with these variables as is we might run into some problems like for example if the gradient is low and we are leaving a sparse region the values for both $m_t$ and $v_t$ will be low and will continue that way for the rest of the iteration, to correct this Bias Correction was introduced in the original paper, which would counteract this situation. Now, to prove for Bias Correction the authors assumed a very strong assumption that all the gradients at every time step have the same distribution hence $\mathbb E [g_i]= \mathbb E[g]$. With bias correction the new $\hat{m_t}$ and $\hat{v_t}$ are defined as,
 
 \begin{equation}
 \begin{split}
 \hat{m_t} = \frac{m_t}{1-{\beta_1}^t}\\
 \hat{v_t} = \frac{v_t}{1-{\beta_2}^t}
 \end{split}
 \end{equation}
 
 And the final update is given by,
 
 \begin{equation}
 x_t= x_{t-1} - \eta \frac{\hat{m_t}}{\sqrt{\hat{v_t}}+\epsilon}
 \end{equation}
 
 Here the ratio of $\hat{m_t}/ \sqrt{\hat{v_t}}$ was termed as Signal to Noise ratio and intuitively speaking as this ratio becomes smaller and smaller so with the effective step size as when the ratio is smaller there is increasing uncertainty if the direction $\hat{m_t}$ is same as the direction of the true gradient and with this, the value will tend to zero when we approach an optimum.
 \subsection{AdaMax}
Adamax was introduced in accordance with Adam and had a similar idea but with only a simple but a counter-intuitive change. As we have been working with Adam we were using the $L_2$ norm hence all the updates and parameters were scaled to $L_2$ norm, we can still generalize for  $L_p$ norm where the stepsize update is now proportional to ${v_t}^{1/p}$ and both $\beta_1$ and $\beta_2$ are in powers of $p$. Hence the new $m_t$ and $v_t$ are,

\begin{equation}
\begin{split}
m_t=\beta_1^p m_{t-1} + (1-\beta_1^p) g_t\\
v_t=\beta_2^p v_{t-1} + (1-\beta_2^p) |g_t|^p
\end{split}
\end{equation}

Now, as $p$ increases harder will be to compute and trace the gradients as we compute them but as the authors found as $p \rightarrow \infty$ we get a stable and a descriable result and the transformation lead to the AdaMax algorithm. As the step size depends on the ${v_t}^{1/p}$ we need to find the limit as $p \rightarrow \infty$, which is given by, $u_t= \lim_{p \rightarrow \infty} (v_t)^{\frac{1}{p}}$ and can be written as $max(\beta_2^{t-1}|g_1|,\beta_2^{t-2}|g_2|,...)$ after solving for the limit in $v_t$. And can be written as a recursive formula as $u_t=max(\beta_2 u_{t-1},|g_t|)$ which is the new $v_t$. And finally the new update will become,
\begin{equation}
x_t= x_{t-1} -  \frac{\eta}{1-\beta_1^t}\frac{m_t}{u_t}
\end{equation}
\subsection{Nadam}
As the name suggests Nadam is the integration of Nesterov’s accelerated Momentum and Adam and was collectively called Nesterov-accelerated adaptive moment\cite{dozat2016incorporating}. Now to use Adam with NAG, we have to modify NAG a little bit to work with the bias correction and the symbology of Adam. We can redefine NAG in the following sense,

\begin{equation}
\begin{split}
g_t=\nabla_{x_t}f(x_{t-1})\\
m_t= \beta_{1^t} m_{t-1} +  g_t\\
\hat{m_t}= \_{t+1} m_{t} +  g_t\\
x_{t+1}=x_t - \eta \hat{m_t}
\end{split}
\end{equation}

Where $\mu_t$ is the same as $\frac{t_k -1}{t_{k+1}}$ at a time step $k$ as defined in subsection C. An important observation in this variation is that the final moment term has the gradient term of the current time step and hence there is no need to calculate the gradient twice. Now, after redefining NAG it is also important to note that the intuition behind it remains the same, this version just improves definitions of the concepts mentioned in the previous subsection. Next, when we look at the Adam optimizer and especially at the final update that is given by equation 46 we know that $\hat{m_t}$ and $\hat{v_t}$ are the bias-corrected terms for sake of solving we are assuming they are not although the final results do include the bias correction as well. So now we can substitute the newly calculated $\hat{m_t}$ in equation 46,

\begin{equation}
x_t=x_{t-1}-\eta \bigg[\frac{\beta_1 m_t}{\Psi} - \frac{(1-\beta_1) g_t}{\Psi}  \bigg]
\end{equation}

Where $\Psi$ is given by ($\Psi= \sqrt{\beta_2 \hat{v_t}+(1-\beta_2){g_t}^2}+\epsilon$). Now an assumption that the original suthour took was that $v_t \approx v_{t-1}$ due to the fact that $\beta_2$ is initailized as $>0.9$ hence we can change the denominator as $\sqrt{v_{t-1}}+\epsilon$ hence when we replace for the denominator of both the terms we get,

\begin{equation}
x_t=x_{t-1}-\eta \bigg[\frac{\beta_1 m_t}{\sqrt{\hat{v_t}+\epsilon}}\frac{(1-\beta_1)g_t}{\sqrt{\hat{v_t}+\epsilon}}  \bigg]
\end{equation}

Now for the sake of readability, we can introduce a new variable $\hat{m_t}$ which is defined by,

\begin{equation}
\hat{m_t}=(1-\beta_1)g_t+\beta_1 m_t
\end{equation}

And the final update becomes,
\begin{equation}
x_t=x_{t-1}-\eta \frac{\hat{m_t}}{\sqrt{\hat{v_t}+\epsilon}}
\end{equation}
\subsection{AMSGrad}
Before we get into the workings of AMSGrad or any further optimizer we have to follow a different set of notations. Let $\mathcal{F}$ be the finite space containing the set of parameters such that $x_t \in \mathcal{F}$. And as defined previously let $f_t$ be the loss function at time step $t$ and the incurred loss at time step will be $f_t(x_t)$. Now due to the uncertainty of the subspace of the cost function at each time step, we have to evaluate using regret which is given by $R(T)=\sum_{t=1}^{T}[f_t(x_t)-f_t(x^*)]$ where $x^*=argmin_{x \in \mathcal{F}}\sum_{t=1}^{T}f_t(x)$. In case of Adam, the regret bound is of the order $O(\sqrt{T})$.  Although we have been using the notion of the final update we haven’t specified the subspace that it belongs to $\mathcal{F}$ hence we define a new final update that projects the final update to the set $\mathcal{F}$ ie., $x_{t+1}=\Pi_{\mathcal{F}}(x_{t-1}-\eta g_t)$. Also as another measure we define $\Gamma_{t+1}$ as the inverse of $\eta$ in adaptive methods and is given by $\Gamma_{t+1}=\bigg(\frac{\sqrt{\hat{v_{t+1}}}}{\eta_{t+1}}-\frac{\sqrt{\hat{v_t}}}{\eta_t}\bigg)$ which is quite important in sense of measuring the step sizes  in each iteration.
\\
\\
AMSGrad \cite{reddi2019convergence} was essentially an improvement on Adam in that sense that the original authors found out that we cannot get a good convergence on adaptive methods with only a limited history of the gradients and they even proved for Adam optimizer. Hence to have a meaningful convergence the optimizer should support the storage of past gradients rather than expectations or moving averages. Another thing to notice is that the learning rates for SGD and ADAGrad can only be decremented that is after initializing the step size can only be reduced or $\Gamma_t \succcurlyeq 0$ for every timestep $t$, but for Adam and RMSprop the values for $\Gamma_{t}$ can be indefinite, which leads to bad convergence for the both. The authors also proved that for $\beta_1,\beta_2 \in [0,1]$ such that $\beta_1/\sqrt{\beta_2} < 1$ Adam doesn’t converge to the optimal solution in a stochastic setting, and even the regret $R$ hence the focus for AMSGrad was to develop an algorithm that has similar implications as Adam that it has bias corrections and adaptive gradients but it should converge to the optimal solution. As mentioned earlier that $\Gamma_t$ in case of Adam is indefinite but was taken as an assumption that it is positive semidefinite. A correction for the step sizes that were introduced with gamma correction was that firstly the step size was initialized as a comparatively smaller quantity as compared to Adam but AMSGrad still has the running notion of suppressing the effect of gradient given that $\Gamma_t$ is positive semidefinite. Another difference between Adam and AMSGrad is that incase of AMSGrad we use the maximum value for $v_t$ for moving averages of the gradient instead of $v_t$ itself hence we can represent the modification as, 
\begin{equation}
\hat{v_t}=max(\hat{v_{t-1}},v_t)
\end{equation}
Hence this actually shows that is the gradients are positive instead of taking a large step size as in the case of Adam, AMSGrad decreases the value for $v_t$ which checks the learning rates even if the gradients are large. And the final update similar to equation is given by,

\begin{equation}
x_{t+1}= \Pi_{\mathcal{F}}\bigg(x_{t} - \eta_t \frac{\hat{m_t}}{\sqrt{\hat{v_t}}+\epsilon}\bigg)
\end{equation}

\subsection{Padam}
Padam\cite{chen2018closing} was an extension of AMSGrad which gave an option to switch between SGD and AMSGrad by introducing a new parameter $p$ whose values lie in the range of $[0,1/2]$, where at $0$ the algorithm is same as SGD with Momentum and at $1/2$ it is similar to AMSGrad. The parameter $p$ was introduced at the final update term for AMSGrad given by,

\begin{equation}
x_{t+1}= \Pi_{\mathcal{F},\hat{v_t}^p}\bigg(x_{t} - \eta_t \frac{\hat{m_t}}{\hat{v_t}^p}\bigg)
\end{equation}

Now, this technique is quite similar to what we saw for AdaMax optimizer as shown by equation 46 but the only key difference is that here we are only changing the final step instead of the norm as above.  Now choosing the value $p$ is the most important task when it comes to this optimizer, but the value of $p$ is highly sensitive to the value of the step size, yet the most promising results were from $p=0.125$. This is quite justifiable as if the value for $p$ was large initially, the value for $\eta/\hat{v_t}^p$ would also large for small $\hat{v_t}$ and this can be resolved easily by using small values for $\eta$. The authors also give a notion of early stopping due to the much decay in the learning rate terming it “small learning rate dilemma” and formulated that as the value for $p$ becomes larger and larger(constrained under [0,1]) a greater effect of the dilemma is observed and if $p<1/2 $ then Padam performs as well, even if the step size is changed a bit. An important note here is that the authors have focussed on improving $\hat{v_t}$ instead of $\eta_t$ because the concept of generality as we strive for better not for a singular problem but we need results that can be applied to a variety of problems, this loss of generality is the reason why optimizers like Adam don’t give remarkably better results than SGD with momentum\cite{gastaldi2017shake} on datasets like CIFAR-10 and CIFAR-100 \cite{krizhevsky2009learning}

\subsection{AdamW}
After introducing a good foundation of gradient descent optimization, there is quite a strong connection between concepts of loss functions and optimization. As introduced in section 2 the loss functions have a regularization term which is called the $\mathcal{L}_2$ regularization term, which exists to prevent overfitting. As a prerequisite, we assume that the loss function$f_t(x)$ includes the regularization term. Another notion of weight decay although not introduced earlier, is just adding a decay term to the weights hence it is given by,

\begin{equation}
x_{t+1}=(1-\lambda)x_t-\eta\nabla f_t(x_t)
\end{equation}

Here $\lambda$ is the rate at which decay is happening. Although these terms $\mathcal{L}_2$ regularization and weight decay are used interchangeably which is not correct. Although while working with SGD with Momentum we can assume that these are the same. For representation, we will define a new representation for the regularized loss functions as,

\begin{equation}
f_t’(x)=f_t(x)+\frac{\lambda’}{2}||x||_2^2
\end{equation}

Where again $\lambda$ defines the weight decay. And now if we substitute equation 57 in equation 58 we will get,

\begin{equation}
\begin{split}
x_{t+1}= x_t -\eta \nabla f_t’(x_t)\\
x_{t+1}= x_t -\eta \nabla f_t(x_t)-\eta \lambda’x_t\\
x_{t+1}=(1-\lambda)x_t-\eta\nabla f_t(x_t)
\end{split}
\end{equation}

Here the only assumption is to take $\lambda’=\lambda/\eta$. This is probably the reason why both methods are mentioned interchangeably. But with this also comes this notion that $\lambda’$ and $\eta$ are strongly coupled which shouldn’t be the case as these hyperparameters are responsible for two separate phenomena. Hence in SGDW, the authors proposed that the weight decay and gradient update to be done simultaneously which is worked by introducing a new variable $\delta_t$ which can be fixed or can have the decay effect as well, hence the update is given by 

\begin{equation}
x_{t+1}=x_t - m_t -\delta_t\eta x_t
\end{equation}

Now, as we proved that for SGDW that $\mathcal{L}_2$ regularization is the same as weight decay, this is not the case for adaptive methods. Let us define a new update for generic adaptive methods(without weight decay) which is given by $x_{t+1}=(1-\lambda)x_t-\eta M_t \nabla f_t(x_t)$ where $M_t$ is a  called a preconditioner matrix and is defined as $M_t \neq kI$ and it can be assumned that it holds the ratio of $\hat{m_t}/ \sqrt{\hat{v_t}}$ also the update with weight decay can be written $x_{t+1}=(1-\lambda)x_t-\eta M_t \nabla f_t(x_t)$  so if we follow the previous notion as given by equation 58 we get,
\begin{equation}
\begin{split}
x_{t+1}= x_t -\eta M_t\nabla f_t(x_t)-\eta M_t \lambda’x_t\\
x_{t+1}=(1-\lambda)x_t-\eta M_t \nabla f_t(x_t)
\end{split}
\end{equation}

Now for the condition to hold that $\mathcal{L}_2$-regularization is the same as weighed decay even for adaptive methods both subequations under equation 59 should be equal that is $\lambda x_t = \eta \lambda’ M_t x_t$ should hold good, but for this to be satisfied $M_t$ has to be a diagonal matrix, which it is not hence this relation doesn’t hold good. Hence for adaptive methods, $\mathcal{L}_2$ is not the same as weighted decay. And due to this reason, we have to make changes to Adam by introducing a weighted decay term. As we did for SGD with momentum even for adam, the final update now becomes, 5

\begin{equation}
x_t= x_{t-1} - \delta_t \bigg[\eta\frac{\hat{m_t}}{\sqrt{\hat{v_t}}+\epsilon}+ \lambda x_{t-1} \bigg]
\end{equation}

\section{Conclusion}
The conclusion goes here.


\section*{Acknowledgment}
\bibliographystyle{ieeetran}
\bibliography{References}

\begin{thebibliography}{10}
\providecommand{\url}[1]{#1}
\csname url@samestyle\endcsname
\providecommand{\newblock}{\relax}
\providecommand{\bibinfo}[2]{#2}
\providecommand{\BIBentrySTDinterwordspacing}{\spaceskip=0pt\relax}
\providecommand{\BIBentryALTinterwordstretchfactor}{4}
\providecommand{\BIBentryALTinterwordspacing}{\spaceskip=\fontdimen2\font plus
\BIBentryALTinterwordstretchfactor\fontdimen3\font minus
  \fontdimen4\font\relax}
\providecommand{\BIBforeignlanguage}[2]{{%
\expandafter\ifx\csname l@#1\endcsname\relax
\typeout{** WARNING: IEEEtran.bst: No hyphenation pattern has been}%
\typeout{** loaded for the language `#1'. Using the pattern for}%
\typeout{** the default language instead.}%
\else
\language=\csname l@#1\endcsname
\fi
#2}}
\providecommand{\BIBdecl}{\relax}
\BIBdecl

\bibitem{nie2018investigation}
F.~Nie, Z.~Hu, and X.~Li, ``An investigation for loss functions widely used in
  machine learning,'' \emph{Communications in Information and Systems},
  vol.~18, no.~1, pp. 37--52, 2018.

\bibitem{ghosh2017robust}
A.~Ghosh, H.~Kumar, and P.~Sastry, ``Robust loss functions under label noise
  for deep neural networks,'' in \emph{Thirty-First AAAI Conference on
  Artificial Intelligence}, 2017.

\bibitem{zhao2015stochastic}
P.~Zhao and T.~Zhang, ``Stochastic optimization with importance sampling for
  regularized loss minimization,'' in \emph{international conference on machine
  learning}, 2015, pp. 1--9.

\bibitem{warren2018transcendental}
M.~Warren, J.~Gresham, and B.~Wyatt, ``Transcendental functions with a complex
  twist,'' \emph{arXiv preprint arXiv:1805.05320}, 2018.

\bibitem{huber2004robust}
P.~J. Huber, \emph{Robust statistics}.\hskip 1em plus 0.5em minus 0.4em\relax
  John Wiley \& Sons, 2004, vol. 523.

\bibitem{shrivastava2016training}
A.~Shrivastava, A.~Gupta, and R.~Girshick, ``Training region-based object
  detectors with online hard example mining,'' in \emph{Proceedings of the IEEE
  conference on computer vision and pattern recognition}, 2016, pp. 761--769.

\bibitem{bengio2009curriculum}
Y.~Bengio, J.~Louradour, R.~Collobert, and J.~Weston, ``Curriculum learning,''
  in \emph{Proceedings of the 26th annual international conference on machine
  learning}, 2009, pp. 41--48.

\bibitem{yuille2003concave}
A.~L. Yuille and A.~Rangarajan, ``The concave-convex procedure,'' \emph{Neural
  computation}, vol.~15, no.~4, pp. 915--936, 2003.

\bibitem{kumar2010self}
M.~P. Kumar, B.~Packer, and D.~Koller, ``Self-paced learning for latent
  variable models,'' in \emph{Advances in Neural Information Processing
  Systems}, 2010, pp. 1189--1197.

\bibitem{amari1999improving}
S.-i. Amari and S.~Wu, ``Improving support vector machine classifiers by
  modifying kernel functions,'' \emph{Neural Networks}, vol.~12, no.~6, pp.
  783--789, 1999.

\bibitem{rumelhart1986learning}
D.~E. Rumelhart, G.~E. Hinton, and R.~J. Williams, ``Learning representations
  by back-propagating errors,'' \emph{nature}, vol. 323, no. 6088, pp.
  533--536, 1986.

\bibitem{sun2016complete}
J.~Sun, Q.~Qu, and J.~Wright, ``Complete dictionary recovery over the sphere i:
  Overview and the geometric picture,'' \emph{IEEE Transactions on Information
  Theory}, vol.~63, no.~2, pp. 853--884, 2016.

\bibitem{ge2016matrix}
R.~Ge, J.~D. Lee, and T.~Ma, ``Matrix completion has no spurious local
  minimum,'' in \emph{Advances in Neural Information Processing Systems}, 2016,
  pp. 2973--2981.

\bibitem{ge2015escaping}
R.~Ge, F.~Huang, C.~Jin, and Y.~Yuan, ``Escaping from saddle points—online
  stochastic gradient for tensor decomposition,'' in \emph{Conference on
  Learning Theory}, 2015, pp. 797--842.

\bibitem{kawaguchi2016deep}
K.~Kawaguchi, ``Deep learning without poor local minima,'' in \emph{Advances in
  neural information processing systems}, 2016, pp. 586--594.

\bibitem{choromanska2015loss}
A.~Choromanska, M.~Henaff, M.~Mathieu, G.~B. Arous, and Y.~LeCun, ``The loss
  surfaces of multilayer networks,'' in \emph{Artificial intelligence and
  statistics}, 2015, pp. 192--204.

\bibitem{jin2017accelerated}
C.~Jin, P.~Netrapalli, and M.~I. Jordan, ``Accelerated gradient descent escapes
  saddle points faster than gradient descent,'' \emph{arXiv preprint
  arXiv:1711.10456}, 2017.

\bibitem{anandkumar2016efficient}
A.~Anandkumar and R.~Ge, ``Efficient approaches for escaping higher order
  saddle points in non-convex optimization,'' in \emph{Conference on learning
  theory}, 2016, pp. 81--102.

\bibitem{allen2018natasha}
Z.~Allen-Zhu, ``Natasha 2: Faster non-convex optimization than sgd,'' in
  \emph{Advances in neural information processing systems}, 2018, pp.
  2675--2686.

\bibitem{liang2018adding}
S.~Liang, R.~Sun, J.~D. Lee, and R.~Srikant, ``Adding one neuron can eliminate
  all bad local minima,'' in \emph{Advances in Neural Information Processing
  Systems}, 2018, pp. 4350--4360.

\bibitem{kawaguchi2019elimination}
K.~Kawaguchi and L.~P. Kaelbling, ``Elimination of all bad local minima in deep
  learning,'' \emph{arXiv preprint arXiv:1901.00279}, 2019.

\bibitem{auer1996exponentially}
P.~Auer, M.~Herbster, and M.~K. Warmuth, ``Exponentially many local minima for
  single neurons,'' in \emph{Advances in neural information processing
  systems}, 1996, pp. 316--322.

\bibitem{nocedal2006numerical}
J.~Nocedal and S.~Wright, \emph{Numerical optimization}.\hskip 1em plus 0.5em
  minus 0.4em\relax Springer Science \& Business Media, 2006.

\bibitem{yuan2010line}
G.~Yuan, S.~Lu, Z.~Wei \emph{et~al.}, ``A line search algorithm for
  unconstrained optimization,'' \emph{Journal of Software Engineering and
  Applications}, vol.~3, no.~05, p. 503, 2010.

\bibitem{li2014efficient}
M.~Li, T.~Zhang, Y.~Chen, and A.~J. Smola, ``Efficient mini-batch training for
  stochastic optimization,'' in \emph{Proceedings of the 20th ACM SIGKDD
  international conference on Knowledge discovery and data mining}, 2014, pp.
  661--670.

\bibitem{musso2020stochastic}
D.~Musso, ``Stochastic gradient descent with random learning rate,''
  \emph{arXiv preprint arXiv:2003.06926}, 2020.

\bibitem{caruana2001overfitting}
R.~Caruana, S.~Lawrence, and C.~L. Giles, ``Overfitting in neural nets:
  Backpropagation, conjugate gradient, and early stopping,'' in \emph{Advances
  in neural information processing systems}, 2001, pp. 402--408.

\bibitem{qian1999momentum}
N.~Qian, ``On the momentum term in gradient descent learning algorithms,''
  \emph{Neural networks}, vol.~12, no.~1, pp. 145--151, 1999.

\bibitem{nesterov1983method}
Y.~E. Nesterov, ``A method for solving the convex programming problem with
  convergence rate o (1/k\^{} 2),'' in \emph{Dokl. akad. nauk Sssr}, vol. 269,
  1983, pp. 543--547.

\bibitem{johnson2013accelerating}
R.~Johnson and T.~Zhang, ``Accelerating stochastic gradient descent using
  predictive variance reduction,'' in \emph{Advances in neural information
  processing systems}, 2013, pp. 315--323.

\bibitem{allen2017katyusha}
Z.~Allen-Zhu, ``Katyusha: The first direct acceleration of stochastic gradient
  methods,'' \emph{The Journal of Machine Learning Research}, vol.~18, no.~1,
  pp. 8194--8244, 2017.

\bibitem{duchi2011adaptive}
J.~Duchi, E.~Hazan, and Y.~Singer, ``Adaptive subgradient methods for online
  learning and stochastic optimization.'' \emph{Journal of machine learning
  research}, vol.~12, no.~7, 2011.

\bibitem{pennington2014glove}
J.~Pennington, R.~Socher, and C.~D. Manning, ``Glove: Global vectors for word
  representation,'' in \emph{Proceedings of the 2014 conference on empirical
  methods in natural language processing (EMNLP)}, 2014, pp. 1532--1543.

\bibitem{zeiler2012adadelta}
M.~D. Zeiler, ``Adadelta: an adaptive learning rate method,'' \emph{arXiv
  preprint arXiv:1212.5701}, 2012.

\bibitem{zhou2018convergence}
D.~Zhou, Y.~Tang, Z.~Yang, Y.~Cao, and Q.~Gu, ``On the convergence of adaptive
  gradient methods for nonconvex optimization,'' \emph{arXiv preprint
  arXiv:1808.05671}, 2018.

\bibitem{kingma2014adam}
D.~P. Kingma and J.~Ba, ``Adam: A method for stochastic optimization,''
  \emph{arXiv preprint arXiv:1412.6980}, 2014.

\bibitem{dozat2016incorporating}
T.~Dozat, ``Incorporating nesterov momentum into adam,'' 2016.

\bibitem{reddi2019convergence}
S.~J. Reddi, S.~Kale, and S.~Kumar, ``On the convergence of adam and beyond,''
  \emph{arXiv preprint arXiv:1904.09237}, 2019.

\bibitem{chen2018closing}
J.~Chen, D.~Zhou, Y.~Tang, Z.~Yang, and Q.~Gu, ``Closing the generalization gap
  of adaptive gradient methods in training deep neural networks,'' \emph{arXiv
  preprint arXiv:1806.06763}, 2018.

\bibitem{gastaldi2017shake}
X.~Gastaldi, ``Shake-shake regularization,'' \emph{arXiv preprint
  arXiv:1705.07485}, 2017.

\bibitem{krizhevsky2009learning}
A.~Krizhevsky, G.~Hinton \emph{et~al.}, ``Learning multiple layers of features
  from tiny images,'' 2009.

\end{thebibliography}

\end{document}